\documentclass{scspaperproc}

\usepackage{latexsym}
\usepackage{graphicx}
\usepackage{mathptmx}
\usepackage{float}
\usepackage{amsmath}
\usepackage{amsfonts}
\usepackage{amssymb}
\usepackage{amsbsy}
\usepackage{amsthm}

\usepackage[pdftex,colorlinks=true,urlcolor=blue,citecolor=black,anchorcolor=black,linkcolor=black,bookmarks=false]{hyperref}

\usepackage{hyphenat}
\newtheoremstyle{scsthe}
{8pt}
{8pt}
{\it}
{}
{\bf}
{.}
{.5em}
{}

\theoremstyle{scsthe}

\usepackage{siunitx}

\begin{document}

%
%

\pagestyle{fancyplain}

\thispagestyle{plain}
\firstPageHead{}

\chead{\fancyplain{}{\itshape\small Jansen, Verreycken, Schenck, Blanquart, Connor, Huebel and Steckel \vspace{8pt}}}

\rhead{}
\cfoot{}
\renewcommand{\headrulewidth}{0pt} 

\makeatletter
\let\@internalcite\cite
\def\cite{\def\@citeseppen{-1000}%
    \def\@cite##1##2{(##1\if@tempswa , ##2\fi)}%
    \def\citeauthoryear##1##2##3{##1 ##3}\@internalcite}
\def\citeNP{\def\@citeseppen{-1000}%
    \def\@cite##1##2{##1\if@tempswa , ##2\fi}%
    \def\citeauthoryear##1##2##3{##1 ##3}\@internalcite}
\def\citeN{\def\@citeseppen{-1000}%
    \def\@cite##1##2{##1\if@tempswa, ##2)\else{}\fi}%
    \def\citeauthoryear##1##2##3{##1 (##3)}\@citedata}
\def\citeA{\def\@citeseppen{-1000}%
    \def\@cite##1##2{(##1\if@tempswa , ##2\fi)}%
    \def\citeauthoryear##1##2##3{##1}\@internalcite}
\def\citeANP{\def\@citeseppen{-1000}%
    \def\@cite##1##2{##1\if@tempswa , ##2\fi}%
    \def\citeauthoryear##1##2##3{##1}\@internalcite}
\def\shortcite{\def\@citeseppen{-1000}%
    \def\@cite##1##2{(##1\if@tempswa , ##2\fi)}%
    \def\citeauthoryear##1##2##3{##2 ##3}\@internalcite}
\def\shortciteNP{\def\@citeseppen{-1000}%
    \def\@cite##1##2{##1\if@tempswa , ##2\fi}%
    \def\citeauthoryear##1##2##3{##2 ##3}\@internalcite}
\def\shortciteN{\def\@citeseppen{-1000}%
    \def\@cite##1##2{##1\if@tempswa, ##2\else{}\fi}%
    \def\citeauthoryear##1##2##3{##2 (##3)}\@citedata}
\def\shortciteA{\def\@citeseppen{-1000}%
    \def\@cite##1##2{(##1\if@tempswa , ##2\fi)}%
    \def\citeauthoryear##1##2##3{##2}\@internalcite}
\def\shortciteANP{\def\@citeseppen{-1000}%
    \def\@cite##1##2{##1\if@tempswa , ##2\fi}%
    \def\citeauthoryear##1##2##3{##2}\@internalcite}
\def\citeyear{\def\@citeseppen{-1000}%
    \def\@cite##1##2{(##1\if@tempswa , ##2\fi)}%
    \def\citeauthoryear##1##2##3{##3}\@citedata}
\def\citeyearNP{\def\@citeseppen{-1000}%
    \def\@cite##1##2{##1\if@tempswa , ##2\fi}%
    \def\citeauthoryear##1##2##3{##3}\@citedata}
%
%
%
\def\@citedata{%
    \@ifnextchar [{\@tempswatrue\@citedatax}%
                  {\@tempswafalse\@citedatax[]}%
}

\def\@citedatax[#1]#2{%
\if@filesw\immediate\write\@auxout{\string\citation{#2}}\fi%
  \def\@citea{}\@cite{\@for\@citeb:=#2\do%
    {\@citea\def\@citea{, }\@ifundefined
       {b@\@citeb}{{\bf ?}%
       \@warning{Citation `\@citeb' on page \thepage \space undefined}}%
{\csname b@\@citeb\endcsname}}}{#1}}%

%
\def\@citex[#1]#2{%
\if@filesw\immediate\write\@auxout{\string\citation{#2}}\fi%
  \def\@citea{}\@cite{\@for\@citeb:=#2\do%
    {\@citea\def\@citea{, }\@ifundefined
       {b@\@citeb}{{\bf ?}%
       \@warning{Citation `\@citeb' on page \thepage \space undefined}}%
{\csname b@\@citeb\endcsname}}}{#1}}%

%
\def\@biblabel#1{}
\makeatother

\newdimen\bibindent
\bibindent=.25in

\def\thebibliography#1{\section*{\refname}\list
   {}{\settowidth\labelwidth{[#1]}
   \leftmargin \bibindent
   \itemindent -\bibindent
   \listparindent \itemindent
	 \itemsep 4pt
   \parsep 0pt
   \usecounter{enumi}}
   \def\newblock{}
   \sloppy
   \sfcode`\.=1000\relax}

\setlength{\baselineskip}{12.7pt}

\def\SCSconferenceacro{ANNSIM'23}

\def\SCSpublicationyear{2023}

\def\SCSconferencedates{May 23-26}

\def\SCSconferencevenue{Mohawk College, ON, CANADA}

\title{Cosys-AirSim: A Real-Time Simulation Framework\\Expanded for Complex Industrial Applications}

\author{
\\
Wouter Jansen\\
Erik Verreycken\\
Anthony Schenck\\ 
Nico Huebel\\ 
Jan Steckel\\ [12pt]
Cosys-Lab (FTI)\\
University of Antwerp\\
AnSyMo/Cosys, Flanders Make vzw\\
Groenenborgerlaan 171, 2020 Antwerp, BELGIUM\\
wouter.jansen@uantwerpen.be\\
\and
\\\\Jean-Edouard Blanquart\\
Connor Verhulst\\\\\\ [12pt]
MotionS\\
Flanders Make vzw\\
Gaston Geenslaan 8, 3001 Leuven, BELGIUM\\
\{jeanedouard.blanquart,\\connor.verhulst\}@flandersmake.be\\
}


\maketitle

\section*{Abstract}
Within academia and industry, there has been a need for expansive simulation frameworks that include model-based simulation of sensors, mobile vehicles, and the environment around them. To this end, the modular, real-time, and open-source AirSim framework has been a popular community-built system that fulfills some of those needs. However, the framework required adding systems to serve some complex industrial applications, including designing and testing new sensor modalities, Simultaneous Localization And Mapping (SLAM), autonomous navigation algorithms, and transfer learning with machine learning models. In this work, we discuss the modification and additions to our open-source version of the AirSim simulation framework, including new sensor modalities, vehicle types, and methods to generate realistic environments with changeable objects procedurally. Furthermore, we show the various applications and use cases the framework can serve. \\ \\
\textbf{Keywords:} sensors, procedural generation, digital twins, transfer learning, open-source.

\section{Introduction} 
\label{sec:introduction}

The development of a model and software-based simulation of a single sensor modality or mobile robotic platform is already a non-trivial task. Creating an entire framework that can simulate various sensor modalities and mobile platform types in complex and realistic environments is even more daunting. Modern-day industrial applications, however, often require some form of flexibility in simulation. These applications include deciding ideal-sensor placement, validating sensor design parameters, testing autonomous robotic algorithms, generating training datasets for neural networks, and much more. Therefore, having a single, modular, and open-source framework allows for tackling the many requirements of such applications across several academic or industrial projects and takes less time to develop. 
Commercial frameworks such as Prescan \shortcite{siemensprescan}, AVxcelerate \shortcite{ansys}, DYNA4 \shortcite{dyna4}, VTD \shortcite{vtd}, DriveWorks \shortcite{driveworks}, rFpro \shortcite{rfpro}, or AURELION \shortcite{aurelion} seem to provide the necessary tools to make these applications work. Most often, these commercial frameworks are designed for ADAS (Advanced Driver-Assistance System)/AD (Autonomous Driving) simulation that can do software-in-the-loop and even hardware-in-the-loop simulation of vehicles within detailed environments. However, to be more accessible to the general public, smaller companies, and academic researchers, open-source frameworks are desired, which can also be used for different application types than ADAS/AD. AirSim \shortcite{Shah2017AirSimVehicles}, LGSVL \shortcite{Rong2020LGSVLDriving}, Webots \shortcite{webots} and CARLA \shortcite{Dosovitskiy2017CARLA:Simulator} are the most commonly referenced at the moment. 
\begin{figure}
\centering
\includegraphics[width=0.65\textwidth]{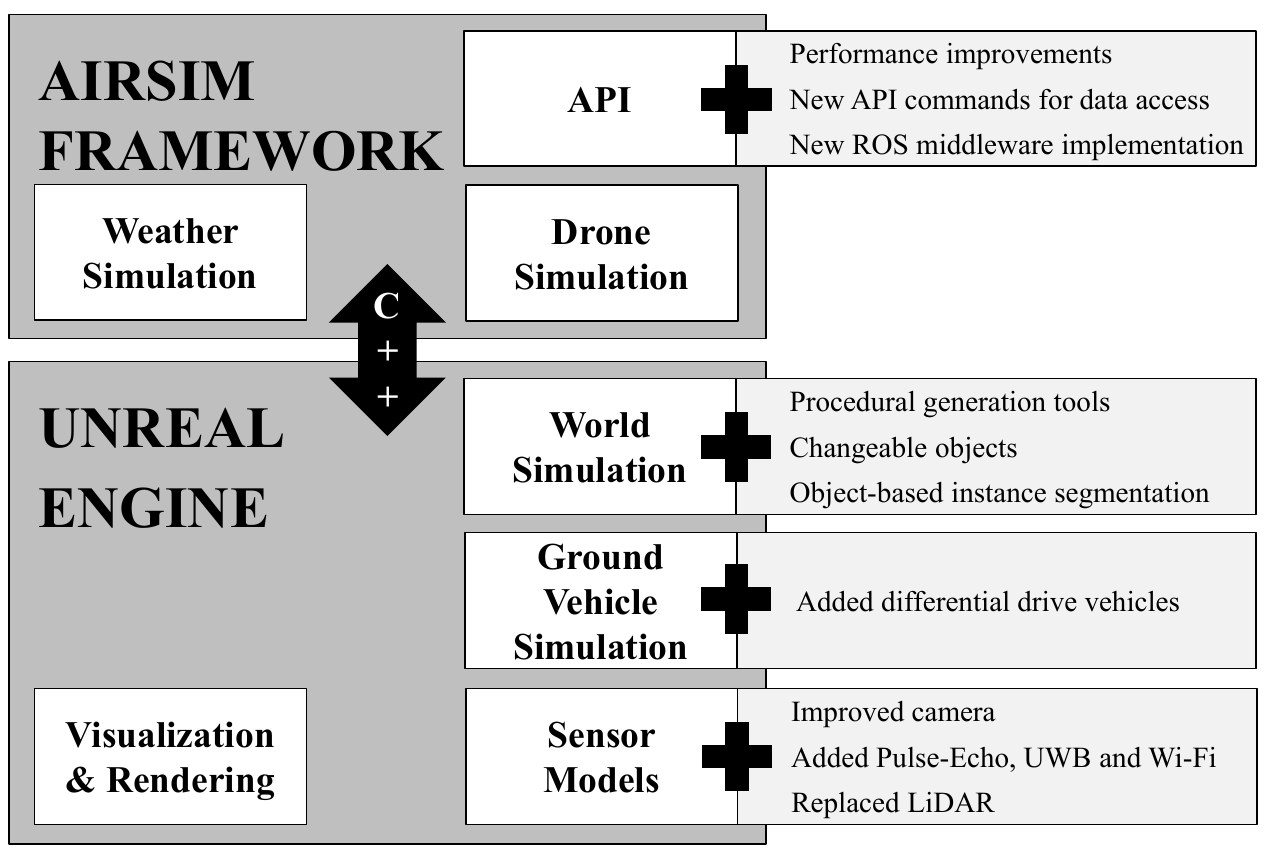}\\
\caption{Block diagram of the software connection between the Unreal Engine and the AirSim Framework and their sub-elements. Furthermore, the major extensions discussed in this paper are listed on the right side.}
\label{fig:airsim_diagram}
\end{figure}
More recently, paradigms such as real-time render engines, particularly those used by video games such as Unreal and Unity, have kick-started the development of the kind of open-source simulation referred to above. When running in real-time, without fixed-step simulation, one has to ensure that the overall compute performance is not hindered by vehicle, sensor, or environmental simulation. Any slowdown could influence other aspects, such as its frame-rate-linked physics and animation systems. Therefore, additional considerations are required to ensure that the desired simulation clock speed is always near real-time. Of the mentioned open-source frameworks, CARLA is built on the Unreal Engine, while LGSVL uses the Unity Engine. AirSim offers both, but the Unreal Engine framework is more stable and fully featured. CARLA is heavily focused on AD car simulation, and the environmental simulation includes map generation and traffic simulation. AirSim and LGSVL offer more alternative mobile platform models, such as drones and indoor robots, respectively. CARLA supports a form of fixed-step simulation, while AirSim and LGSVL only offer real-time simulation with an accelerated or slowed-down clock. Some other frameworks exist, such as Gazebo \cite{Koenig2004DesignSimulator} and MATLAB \cite{mathworksautomation} but do not offer a high level of detail in the environment simulation. AirSim was the preferred platform for tackling several projects within our research group, as it can render high-resolution environments and simulate ground and air-based vehicles. However, various new features and other improvements within AirSim were required for the complex industrial projects within our group. In this paper, we, therefore, present our extensive work on upgrading AirSim with additional sensor modalities and framework improvements. Additionally, while vehicles and sensors are the prominent focus of many simulators to capture high-resolution, realistic sensor data, the environment itself also requires attention. Several techniques were, therefore,  used to facilitate (re)creating realistic, complex, and changeable environments. We discuss our contributions to creating these environments with changeable objects manually or automatically using procedural generation based on a set of rules and constraints. \\ All these contributions can run in the real-time simulation system of AirSim and are available in our open-source repository. Subsequently, we discuss some of the industrial applications we now have the ideal framework for and their use cases. Finally, we conclude the paper with the current state and future plans of the simulation framework. 

\section{Methods}
\label{sec:methods}

\subsection{Simulation Framework Overview}
\label{subsec:framework}

\begin{figure}
\centering
\begin{tabular}{cc}
\includegraphics[width=0.43\textwidth]{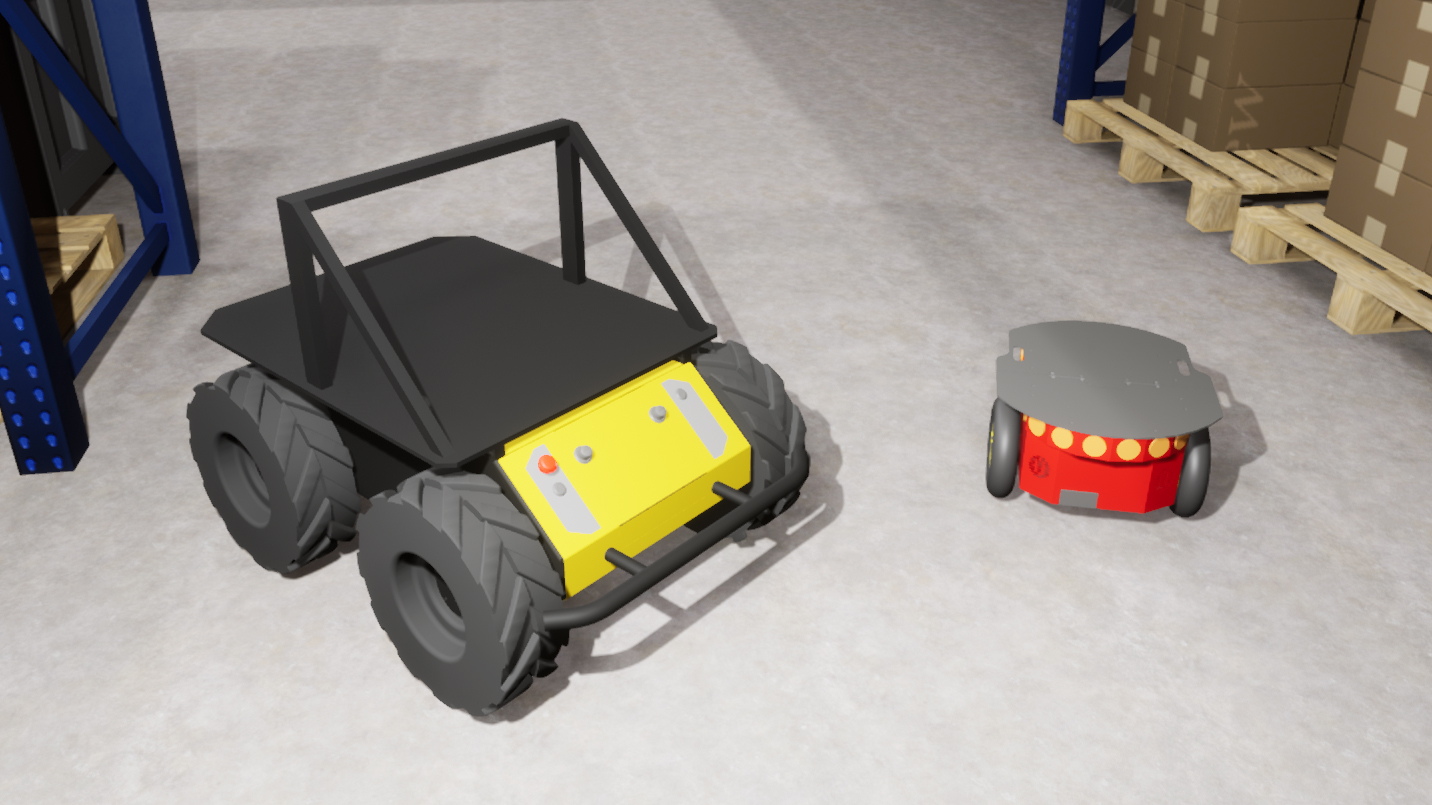}&\includegraphics[width=0.43\textwidth]{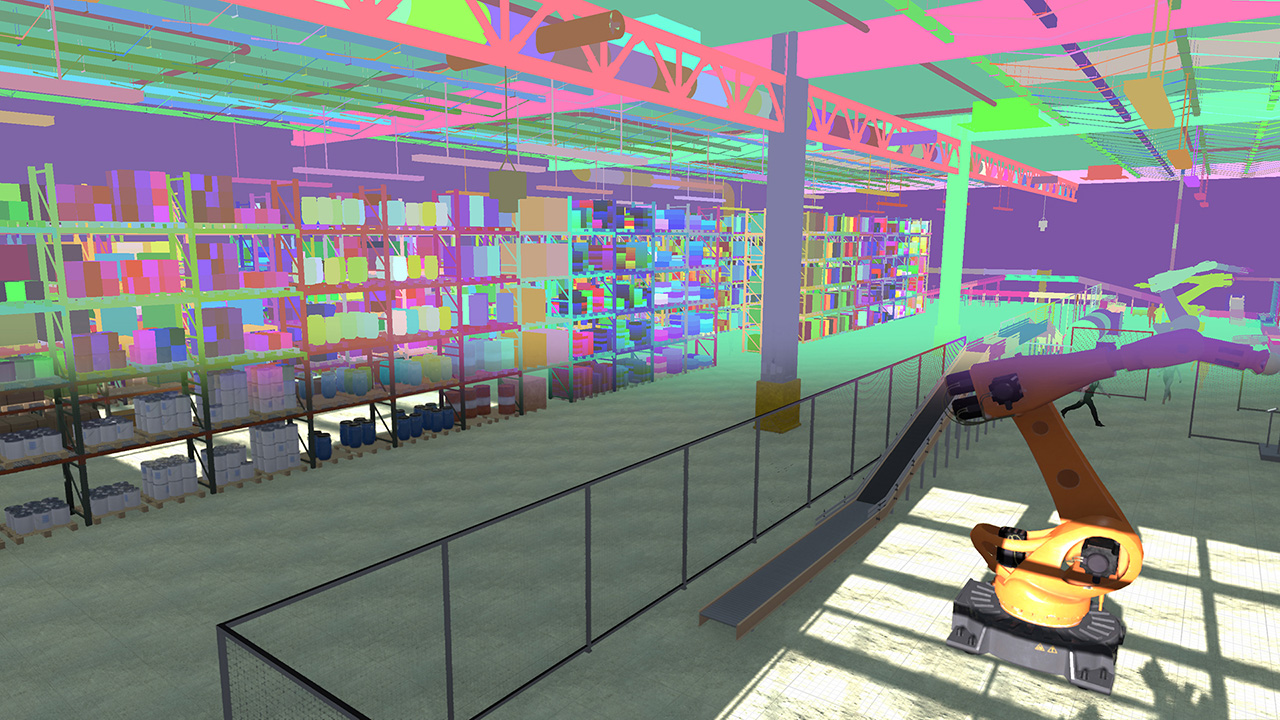}\\
(\textbf{a})&(\textbf{b})\\
\end{tabular}  
\caption{(\textbf{a}) Differential drive steering model vehicles like the Clearpath Robotics Husky and Pioneer P3-DX were added. (\textbf{b}) The unique object-based instance segmentation gives each object a unique label and color.}
\label{fig:airsim_extensions}
\end{figure}

The AirSim framework was initially developed by Microsoft Research in 2017 as an open-source project \shortcite{Shah2017AirSimVehicles}. It was designed for AI research and experimentation. While it focused on Unmanned Aerial Vehicles (UAVs), it was designed to be modular to accommodate new types of mobile platforms, sensors, and environments. Therefore, we chose it as our platform within our group and consequently published our modified version online under the name Cosys-AirSim \cite{Cosys-AirSimRepository}.
In July 2022, Microsoft announced the end of its support to the original AirSim research project: it would evolve into a commercial closed-source version. This type of event further highlights the need for long-term supported, community-driven simulation frameworks. \autoref{fig:airsim_diagram} shows a diagram of the interaction between the AirSim C++ library and the underlying Unreal Engine. It also shows several extensions we made to the framework. \\
One of our extensions to the general framework was a new implementation of the ROS middleware API \shortcite{Quigley2009} which allows sensor and ground truth data extraction; recording and replaying of trajectories for vehicles; and tracking object positions. Because the simulation is running in real-time, the data capturing through the API has to run in real-time, which was achieved. This enables the generation of datasets at a much faster speed than in real life because multiple simulations can run at the same time. 
Our second extension was the addition of differential drive (Autonomous) Ground Vehicles(AGVs) as seen in \autoref{fig:airsim_extensions}a. Lastly, support for object-based labels for all unique instances of those objects to be used for sensor ground truth was added. While AirSim already includes ground truth labels for object classes, it is limited to 255 different object classes. This limits the data generation in, for example, object-detection machine learning (ML) models. In other work, UnrealCV \shortcite{QiuUnrealCV:Engine,Qiu2017UnrealCV:Vision} created a technique to assign a unique color to millions of objects within the environment by using the vertex colors within Unreal Engine. We adjusted this technique for the Cosys-AirSim framework. We upgraded it to the latest versions of the Unreal Engine by making it capable of dealing with gamma-corrected rendering and generating a unique label and RGB888 color that is saved in a data table for each static or runtime-generated object in the environment. The result of these uniquely colored objects in an environment can be seen in \autoref{fig:airsim_extensions}b. Outside of these extensions, minor performance improvements and additional API commands were added. 

\subsection{Sensor Models}
\label{subsec:sensormodels}

The original AirSim suite of sensors contains camera, barometer, IMU, GPS, magnetometer, and Light Detection And Ranging (LiDAR) sensors. We improved the camera model with additional distortion features such as chromatic aberration, motion blur, and lens distortion. However, most efforts were put into completely re-implementing the LiDAR model and adding pulse-echo, UWB (Ultra-Wideband), and Wi-Fi model-based simulation.
Furthermore, we added some new features to all sensor types. Firstly, all sensor types can be placed in a static position in the environment, independent of vehicles. Secondly, objects can be marked as invisible for specific sensors and will not be present in their sensor data streams. This can be beneficial for training specific ML algorithms with specific training data. 

\subsubsection{LiDAR}
\label{subsec:lidar}

 LiDAR sensors can provide a 2D or 3D depth map, usually as a point cloud. The sensor uses the medium of structured light and is subjected to distortions from environmental factors such as weather and the properties of reflecting surfaces. The latter are mainly defined by the angle between the laser beam and the surface, the wavelength of the laser beam, as well as the material of the surface \cite{Muckenhuber2020AutomotiveCapabilities}. Combining all three, the Lambertian target reflectance describes the reflectance measured for the LiDAR laser wavelength for various incidence angles in percentages. This results in a data point associated with a specific laser having a different reflection intensity depending on those three parameters mentioned above. This can also cause it not to be perceived by the sensor if the laser power has attenuated below the sensor's capability. Furthermore, the sensor design properties also influence the final result. Not only the laser distribution has an impact, but, for example, also the reflectance limit function. This limit depicts the maximum range for which a point would be detected depending on the corresponding Lambertian target reflectance, something usually provided in the data sheet of a commercial LiDAR sensor. 
The original simulation of LiDAR sensors in AirSim didn't have a physical model with these characteristics. It used simplistic collision ray tracing calculated sequentially on the CPU, causing it to perform inefficiently for high-resolution LiDAR sensors and not provide physically accurate results. Consequently, we opted to start from scratch and design our own LiDAR simulation model based on the existing work of other peer-reviewed publications. The goal was to provide a real-time simulation of a physically accurate LiDAR model that can deal with different surface materials and sensor properties. In this paper, we provide a synopsis of the implementation within our Cosys-AirSim framework. A more detailed description can be found in \citeN{lidarsim2022jansen}. To perform the simulation in real-time, the implementation uses several GPU-accelerated shaders. The latter are used on a virtual render camera that, for each frame (a single time step of the simulation engine), will be rotated according to how much time has passed as the simulation is real-time and not fixed-step. The resulting camera images are subsequently parsed and turned into point cloud data on the CPU by translating the desired LiDAR laser azimuth and elevation angles to the correct virtual render camera's pixel coordinates. The developed GPU-Accelerated shaders were used for the following: 
\begin{itemize}
    \item Depth calculation: Range estimation for each laser. This comes from the Unreal Engine G-Buffer during the depth render pass. 
    \item Object-based instance segmentation: A unique RGB value is attributed to each unique object using vertex colors as explained in \autoref{subsec:framework}. 
    \item Surface material \& inclination angle determination: The returned laser power is influenced by the surface material and the impact angle at which the LiDAR laser hits the surface of an object. This information is extracted by combining two shaders, one for the material type and one for the inclination angle. For the surface material reflectance values, the NASA ECOSTRESS spectral library \cite{JetPropulsionLaboratoryECOSTRESS1.0} is used.
\end{itemize}
The sensor capability, i.e., the optical performance of the sensor, such as maximum range and reflectance limit function, is implemented on the CPU side during the point cloud generation. \\
Finally, a new contribution to the LiDAR model includes the impact of rain on the LiDAR sensor performance. AirSim already provides basic weather simulation; this system was extended to influence sensors such as LiDAR. We implemented the model by \shortciteN{Goodin2019PredictingADAS} introducing the performance degradation of a LiDAR in rainy conditions regarding the intensity and range measurements. They simplified the LiDAR receive power equation of \shortciteN{Dannheim2014WeatherSystems} and reduced it in function of the fractional reduction, $\delta$ as caused by rain to the power or intensity of a LiDAR to $\delta = e^{-2\alpha d} - 1$. With $d$ being the measured LiDAR distance and $\alpha$ the rain scattering coefficient. \shortciteN{Lewandowski2009Lidar-basedEvidence} derive the relationship between the rain scattering coefficient and the commonly used rainfall rate $R$ as a power law with two empirical coefficients $a$ and $b$ for a 90\% diffusely reflecting surface  to $\alpha = aR^b$.
The rain also modulates the measured range. We integrated the range error model by \shortciteN{Goodin2019PredictingADAS} where the error is sampled from a normal distribution around the true range with a standard deviation $\sigma = 0.02d(1-e^{-R})^2$. 

\subsubsection{Pulse-Echo}
\label{subsec:pulsecho}

Echolocation is the concept of a single or multiple emitter(s) sending a designed pulse signal into the environment, reflecting off objects within it. Some of those reflected signals are captured by single or multiple receivers. These are called echoes, from which, after signal processing, information about the reflected objects, such as their positioning and surface material, can be retrieved. This is done by looking at the timing and intensity of the returned echoes. In nature, we find this principle predominantly in certain bats, whales, and dolphins species. Humans, as always, have copied this principle to create sensors that work similarly. This sensor is called RADAR in the electromagnetic spectrum, while in the acoustic spectrum, it is called SONAR. However, the pulse-echo model was implemented to support all spectra. In this paper, we provide a synopsis of the implementation within our Cosys-AirSim framework. A more detailed description can be found in \citeN{Schouten2021SimulationSLAM}.
The model generates a 3D or 2D point cloud representing a sparse set of echoes received by the pulse-echo sensor via collision ray tracing. This technique creates a sparse recreation of how the signal propagates through free space and interacts with the objects it comes in contact with. If the reflected signal passes through the sensor’s receptive field based on the antenna (receiver) positions, it produces an echo waveform in the resulting measurement. \\ The rays are cast uniformly within the described field of view of the sensor. Each ray travels until it collides with an object. Otherwise, it dissipates. Dissipation is defined by the signal attenuation limitation described by the sensor. A ray's attenuation is determined by free-space path loss and atmospheric attenuation. The weather simulation within AirSim influences this last form of attenuation. If the ray collides with an object, its direction is changed according to the specular reflection characteristics of the sensor. The sensor's receiver is specified by its opening angle, determining if the antenna captures the reflected echo. 

\subsubsection{UWB \& Wi-Fi} 
\label{subsec:uwbwifi}

Ranging sensors based on Radio Frequency (RF) signals such as UWB or Wi-Fi can measure the range from the sensor or tag to a static anchor with a known position. The sensor periodically sends poll signals to which the static anchor can respond. The time between the poll and response, or time of flight, can then be used to calculate the position of the sensor using a localization algorithm such as time difference of arrival.\\
Our implementation uses collision ray tracing similar to the echo sensor to create a sparse representation of the RF signal wave model.  This differs from the echo sensor in how the signal propagates through free space and reflects or attenuates in response to certain material types. Another difference is that these sensors require specific anchors that will be the receivers of the RF signals. They are added at runtime to ensure that the static anchors can be moved without recompilation of the environment. \\This is achieved manually or automatically by adding tags to the environment as described later in \autoref{subsec:proceduralgeneration}. The developed sensor model can predict the paths of rays from the transmitters to the tags or anchors. The model derives the angle of departure, arrival angle, and arrival time. Furthermore, it estimates the path loss and phase change for each ray.
 \subsection{Environments}
\label{sec:environments}
\subsubsection{Realistic Recreation}
\label{subsec:recreation}
\begin{figure}
\centering
\includegraphics[width=1\textwidth]{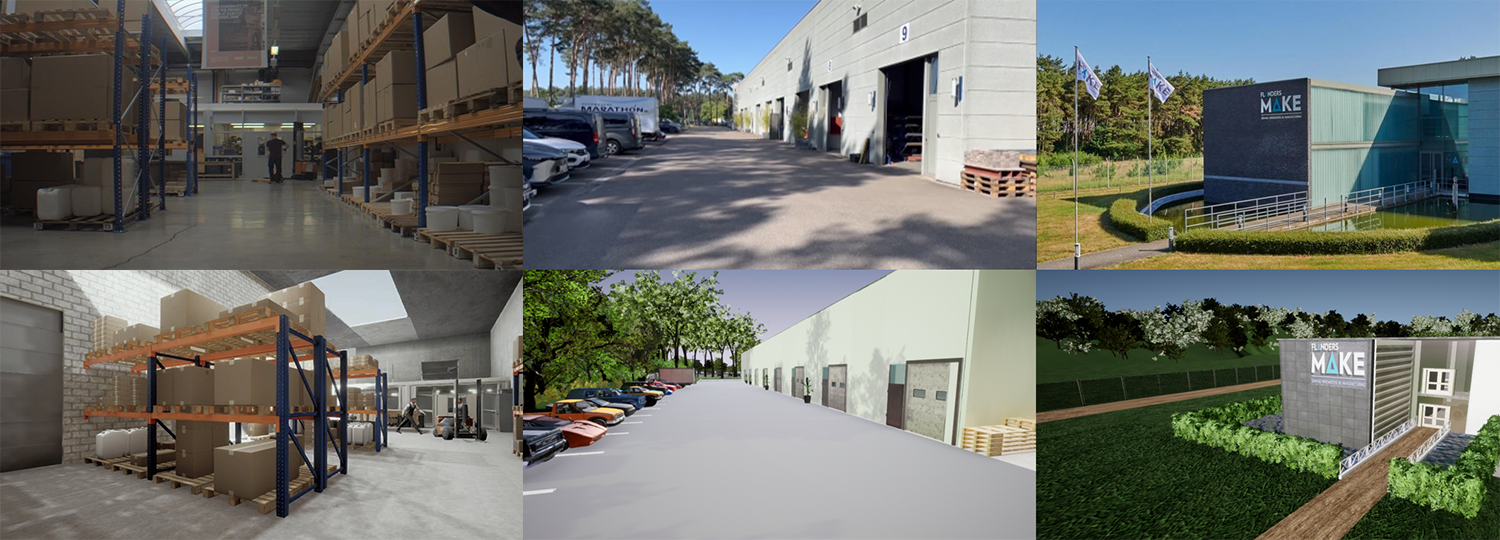}
\caption{The comparison between a real warehouse and outdoor environment on the top row and its counterpart in the simulation shown on the bottom row.}
\label{fig:realisticenvironments}
\end{figure}

Most environments made in simulation can be used to create situations that are hard to produce in real life. These could be environments and conditions that the user typically does not have access to or that could have safety or privacy concerns. However, in other applications, such as ML model transfer learning, the environments must provide high detail and match an existing real location. We re-created some real-life environments within the simulator as seen in the example shown in \autoref{fig:realisticenvironments}. Special attention was paid to lighting conditions and surface materials.
\subsubsection{Changeable Objects}
\label{subsec:changeableobjects}
\begin{figure}
\centering
\includegraphics[width=0.86\textwidth]{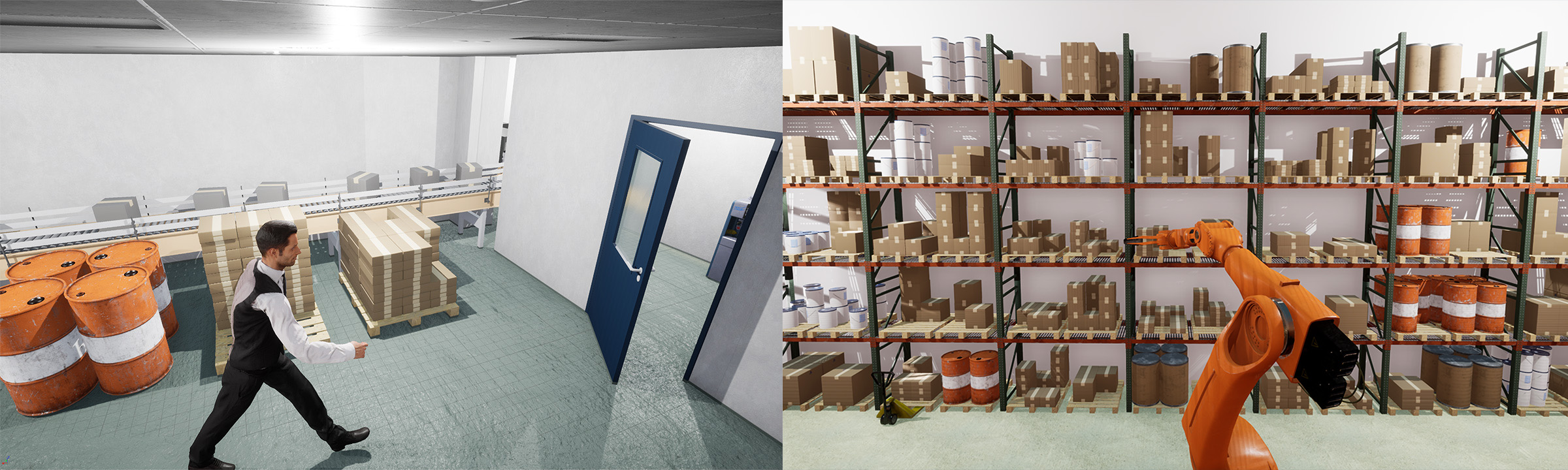}
\caption{Some examples of changeable objects created within our environments include path-following humans, randomized stacked pallets and racks, conveyor belts, a robotic arm, and a door.}
\label{fig:changeableobjects}
\end{figure}

While a static high-resolution and realistic environment is beneficial for applications that only require sensor-data generation, other applications, such as validation of navigation and Simultaneous Localization And Mapping (SLAM) algorithms, require more dynamic and active environmental simulation. We added object models with different states, such as doors and windows that can be open, closed, or something in between. Another example we added is a warehouse environment in which each pallet can be stacked slightly differently or where a pallet rack can have different contents in each simulation run. Also, consumable and human-interactive objects such as coffee cups, food items, brooms, and cones can be randomly placed in the world. Furthermore, multiple moving actors such as humans, forklifts, and other AGVs can be added to navigate between waypoints with simple collision-avoidance behavior or following predetermined splines. Finally, machines such as robotic arms and conveyor belts provide additional environmental motion. 
\newpage All these so-called changeable objects were implemented model-driven and can be customized with several parameters to determine their randomization characteristics. To make the entire system deterministic, all randomization is based on a single seed value that can be manually set or generated using a pseudo-random number generator. Some examples of changeable objects can be seen in \autoref{fig:changeableobjects}.

\subsubsection{Procedural Generation} 
\label{subsec:proceduralgeneration}
\begin{figure}
\centering
\begin{tabular}{cc}
\includegraphics[width=0.3145\textwidth]{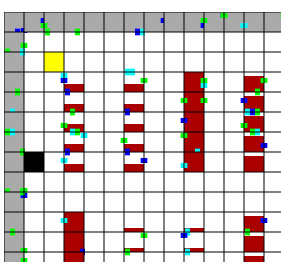}&\includegraphics[width=0.5\textwidth]{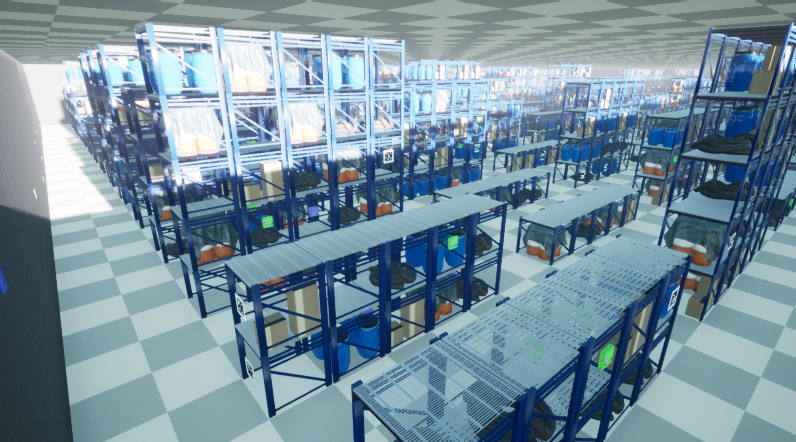}\\
(\textbf{a})&(\textbf{b})\\
\end{tabular}  
\caption{Example of a procedurally generated warehouse environment. (\textbf{a}) shows a simplified view in an external visualizer/editor, where gray is used for walls; red for pallet racks with a height of \SIrange{1}{5}{\meter} indicated by the fill level of the square; blue, green and cyan are used for RF anchors and fiducial markers. \textbf{b}) is the Cosys-AirSim representation of the same environment.}
\label{fig:proceduralgeneration}
\end{figure}

Testing the robustness of a SLAM algorithm requires testing multiple variations of the same environment, which can be done using changeable objects as described in \autoref{subsec:changeableobjects}. However, to test a completely new environment, human interaction is still needed to place (changeable) objects and (re)compile the environment. A completely new method was added to Cosys-AirSim to overcome these limitations. Procedural generations allow the creation of variations (including randomized changeable objects) of real environments or brand-new environments much faster than would be possible in the real world. With procedurally generated environments, a list of objects that must be placed is provided to Cosys-AirSim at runtime. These objects are then added to the environment to create a new variant. Objects that can be placed in this manner are, e.g., walls, pallet racks, pallets with randomized contents, e.a. The list of objects and their locations can be generated by an external tool with additional constraints where the properties of all objects are known. Furthermore, the API can retrieve and use this information in third-party software. \autoref{fig:proceduralgeneration} shows an example of a generated map and a simplified representation of that which can be used in an external learning algorithm.

\section{Applications}
\label{sec:applications}

This section will discuss multiple complex industrial applications for which our updated Cosys-AirSim framework can be used.

\subsection{Ideal Sensor Placement} 
\label{subsec:sensorplacement}
As discussed in \autoref{subsec:uwbwifi}, knowing where to place these sensors and their tags and static anchor receivers is essential. Similarly, fiducial markers such as ArUco or QR codes must be placed for camera sensors. This often depends on the application these sensors need to serve, such as a localization algorithm. This is an ideal use case to use a simulation framework.\\
Generated environments, as described in \autoref{subsec:proceduralgeneration}, can be achieved by a ML algorithm. By constraining what objects can be placed and dividing the environment in a uniform grid, the complete environment can be represented as a single bitmap which can be used as input for a ML algorithm.  \\
The generated environment can then be evaluated, e.g., by calculating the error of a localization algorithm, which can be used as feedback to the ML algorithm. This enables the algorithm to iterate over environments and sensor placement in these environments quickly. This method is used to learn about the relationship between the environment and the rules for placing sensors. Two things can be learned from this approach. Firstly, a prediction about the overall localization given the environment and the placed sensors can be made quickly without testing the localization error at several locations in the environment. Secondly, a prediction can be made about the ideal sensor placement given a particular environment. 

\subsection{Camera Transfer Learning}
\label{subsec:cameralearning}

\begin{figure}
\centering
\begin{tabular}{ccc}
\includegraphics[width=0.3\textwidth]{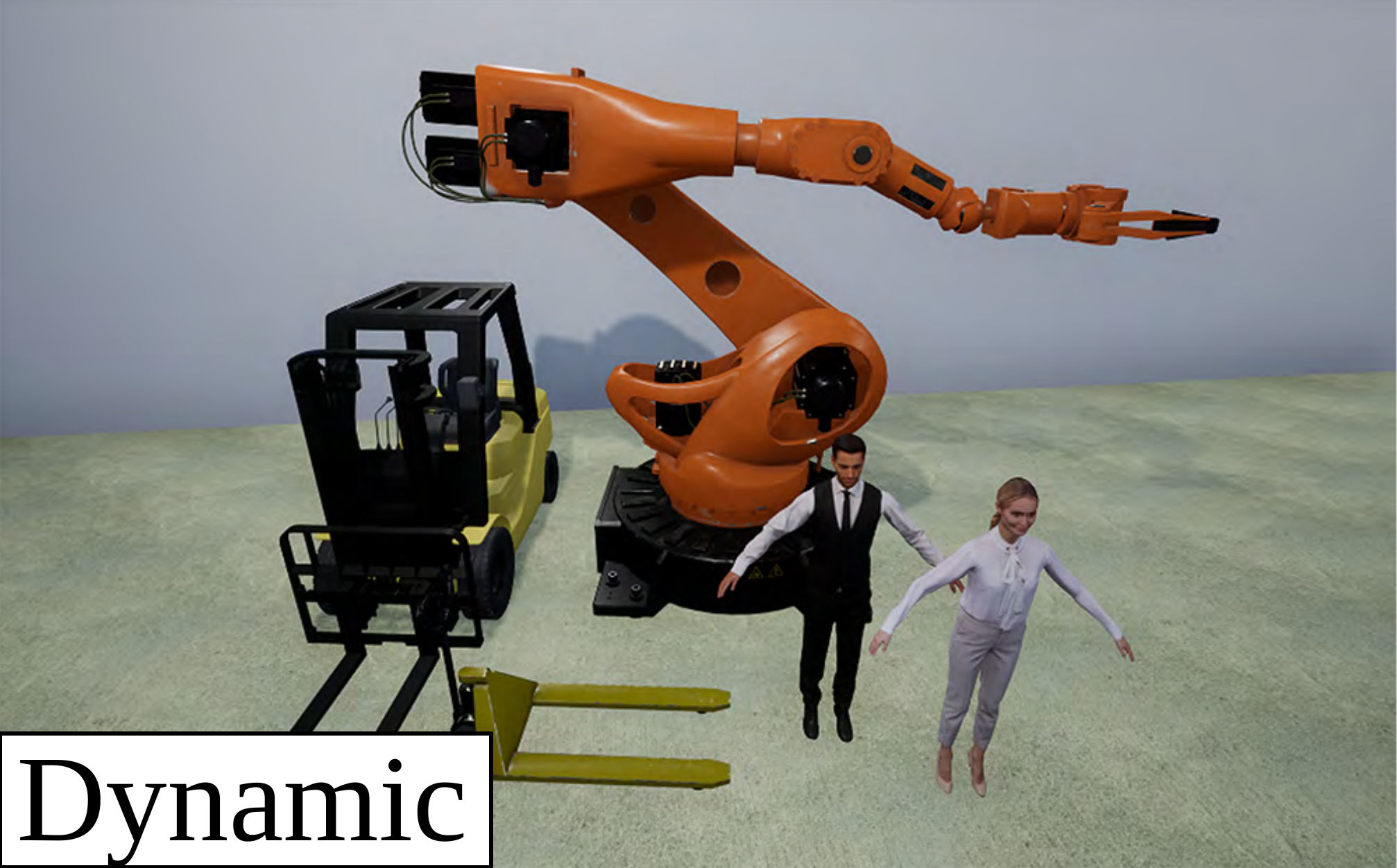}&\includegraphics[width=0.3\textwidth]{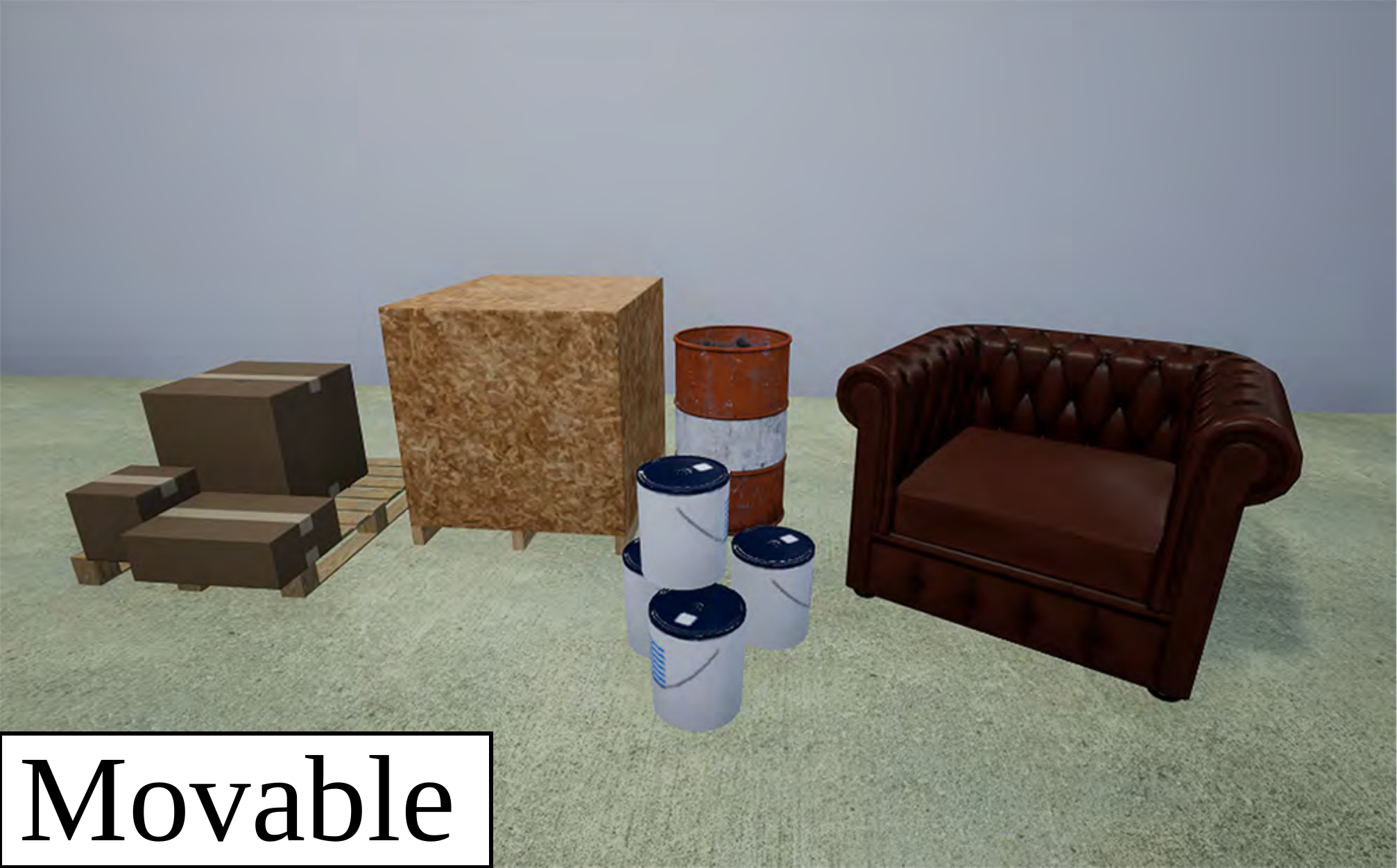}&\includegraphics[width=0.3\textwidth]{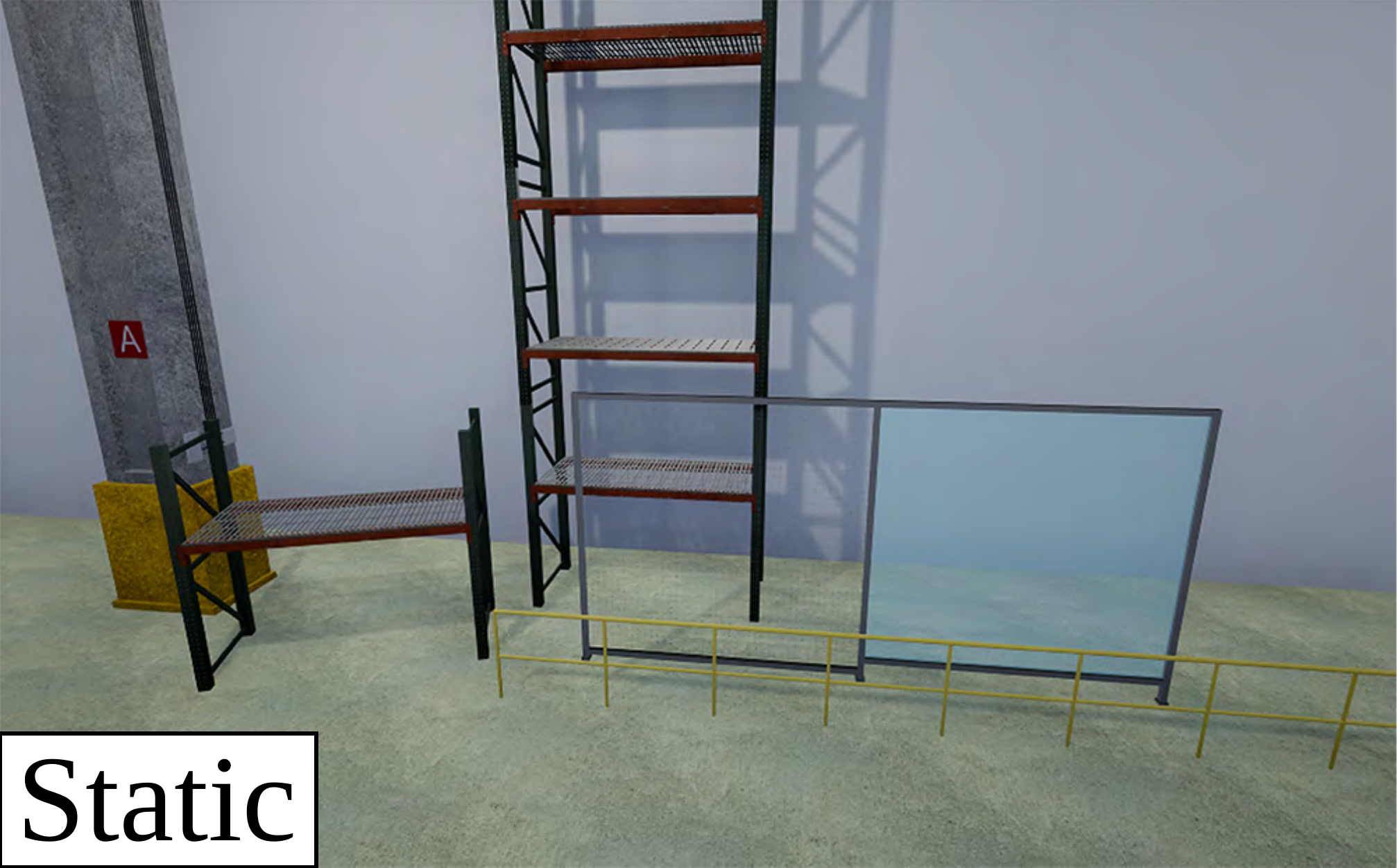}
\end{tabular}  
\caption{Different object categories used in object segmentation of sensor data to improve object-based localization algorithms. By masking out dynamic objects, the navigation will be more stable.}
\label{fig:assettype}
\end{figure}

\begin{figure}
\centering
\includegraphics[width=0.95\textwidth]{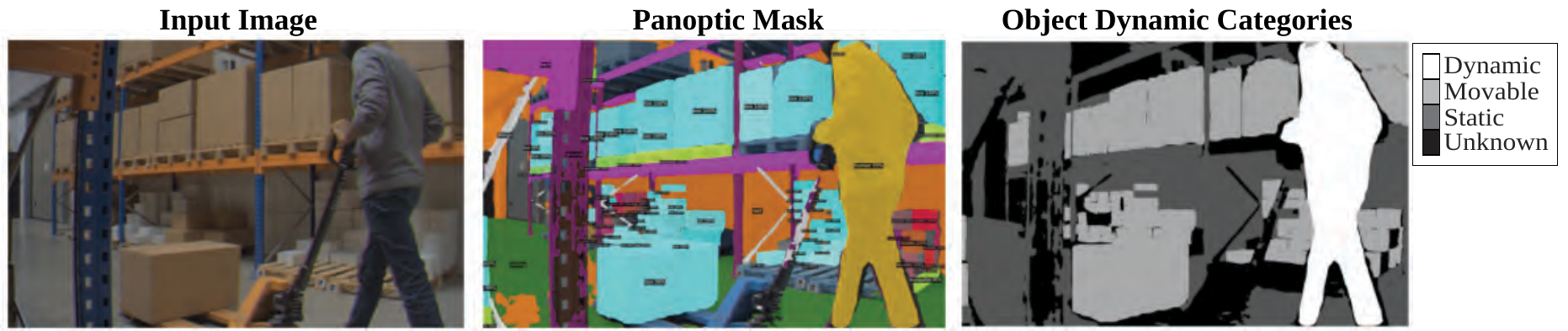}
\caption{Object dynamic state masking with panoptic segmentation.}
\label{fig:panopticsegmentation}
\end{figure}

SLAM algorithms rely on features in the environment to estimate the relative displacement of the camera. However, when those features are moving independently from the camera and the environment, the relative motion can be miss-estimated, resulting in a localization error. To limit this effect, we rely on ML models and, more specifically, panoptic segmentation to identify the objects in the surrounding environment and classify them into three categories: static, movable, and dynamic, as seen in \autoref{fig:assettype}. \\
By gaining insight into the dynamics of the surrounding objects as seen in \autoref{fig:panopticsegmentation}, we can discard features from dynamic objects and apply more weight to movable objects for short-term localization and to static objects for long-term localization, thus improving the robustness of the algorithm. 
However, to deploy this type of algorithm in a changing industrial context, the ML model must often perform well from the get-go. \newpage 
To help achieve this, we developed a solution that mimics the real environment in simulation as discussed in \autoref{subsec:recreation} and seen in \autoref{fig:realisticenvironments}. This synthetic representation of the real world is then used to generate a large amount of synthetic data using the physically accurate camera sensor simulation available in the framework and to pre-train the ML models. A real-time, high-resolution simulator such as Cosys-AirSim enables the generation of large quantities of training data in various environments with different conditions and variations, which would not be possible in the real world in the same time frame. Once the system is deployed, real data can be collected and annotated to improve the model performance further. Thanks to the diversity and quantity of assets available on the Unreal marketplace, building such an environment is easy, cheap, and time efficient. 

\subsection{LiDAR Point Cloud Labeling}
\label{subsec:lidarlearning}

\begin{figure}
\centering
\begin{tabular}{cc}
\includegraphics[width=0.45\textwidth]{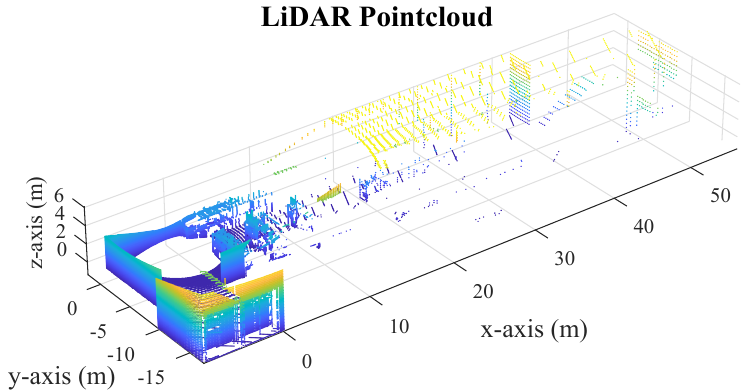}&\includegraphics[width=0.5\textwidth]{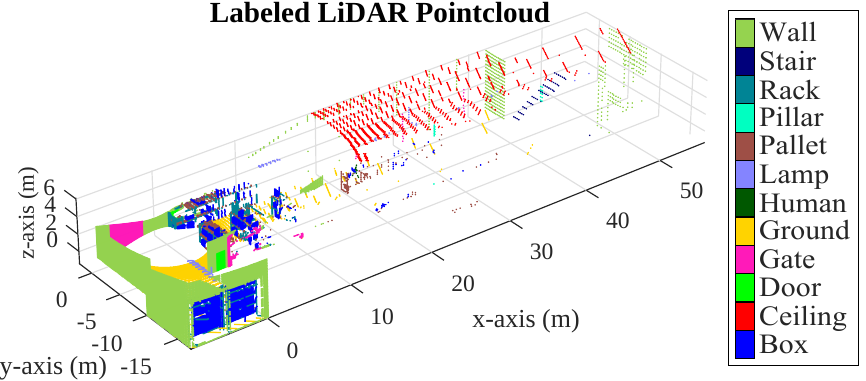}\\
\end{tabular}  
\caption{Example of a single LiDAR scan from the simulator. The data used to create a depth image is shown on the left. The figure on the right shows the labeled point cloud from which the label image is created.}
\label{fig:lidarlabel}
\end{figure}

Current state-of-the-art LiDAR-based SLAM algorithms for autonomous vehicles and robots can encounter problems when used in changing industrial contexts such as warehouses. To increase the performance of these SLAM algorithms, which rely on data generated by a LiDAR sensor, it can be beneficial to filter out data points that belong to dynamic and movable objects. To do this, we need to be able to classify these objects within a LiDAR scan before passing its data on to the SLAM algorithm. This can be done using neural networks. By either first transforming the point cloud to an image to process \cite{Jhaldiyal2022SemanticSO}, or directly on a LiDAR scan \shortcite{qi2017pointnetplusplus,8579077,NEURIPS2018_f5f8590c}.
The Cosys-AirSim framework allows the easy creation of physically accurate labeled LiDAR data sets as discussed in \autoref{subsec:lidar}, which we can use to train a neural network to perform semantic segmentation. This has proven invaluable since there are very few, if any, real-world labeled data sets that focus on indoor environments. This starkly contrasts with outdoor data sets, which have the benefit of autonomous driving research \cite{geiger2012cvpr}. Manual labeling of such a real-world indoor data set is also costly, both monetary and time-wise. In addition, indoor environments are much more diverse, with possible use cases ranging from residential to industrial environments, warehouses, and even greenhouses. Using the simulator, we can quickly create an environment that closely resembles the specific use case and create a data set with which we can train a neural network. An example LiDAR scan and its labeled version can be seen in \autoref{fig:lidarlabel}.

\subsection{Autonomous Navigation and SLAM}
\label{subsec:slam}

While validating autonomous mobile platforms and their algorithms for navigation, localization and mapping in the real world is always required, having safe virtual environments to test is an enormous benefit. \citeN{Schouten2021SimulationSLAM} used the Cosys-AirSim framework to validate if it is sufficiently accurate to be used in developing vehicle controllers and SLAM algorithms for pulse-echo radar sensors as those discussed in \autoref{subsec:pulsecho}. A synthetic representation of a real office environment was recreated in simulation and navigation and SLAM algorithms was compared between a real differential drive AGV with pulse-echo RADAR and their simulation counterparts. \\ The navigation algorithm running in the simulation had an average deviation of ±1.9\% compared to the real-world experiment for the high-priority navigation modes. The SLAM algorithm had an error smaller than \SI{3.2}{\meter} in 95\% of the total trajectory, much lower than the minimum error of \SI{10}{\meter} the odometry-based positioning algorithm had in 85\% of the same trajectory.
The results show strong resemblances, demonstrating that the sim-to-real gap is reduced thanks to the framework's sensor, vehicle, and environmental simulation. This validation allows use cases to develop different RADAR sensor configurations and to test them for various autonomous robotic algorithms within our projects. 

\section{Conclusion and Future works}
\label{sec:conclusion}

The Cosys-AirSim framework has been a stable and extensive set of tools tested in various industrial applications. The challenge of keeping the simulation capable of running in real-time continues to be one of its complex constraints. It might require implementing some form of fixed-step time simulation in the future. Furthermore, future projects might require additional features, such as new sensor models and tools to develop the environments. E.g. currently, a simulation environment based on the real world is a disconnected digital twin. In the near future, we will add the possibility to link the digital objects, vehicles, and other actors through the simulation API based on their real-world counterparts so that they can be moved and adjusted accordingly in the digital twin. Connecting the simulation environment to real-time position-tracking systems to record and replay challenging scenarios would be possible. This would generate synthetic contextual data to augment the training sets enabling fast and efficient retraining of ML models to cope better with these scenarios. Finally, generating UAV training data can be challenging, mainly when a flight permit is necessary or when the environment is a no-fly zone or densely populated area. Therefore, we are currently recreating outdoor environments to generate relevant synthetic data of real outdoor environments and re-train the segmentation models for several sensor models so that they can be used for UAV-capable scenarios. An example of this ongoing work is shown in \autoref{fig:realisticenvironments} in the outdoor environment. By publishing our work open-source, our group and the community can continue to build upon the existing framework and provide the right set of tools for the vast demand for realistic simulation systems. 

\section*{Acknowledgments}
This research was partially supported by Flanders Make, the strategic research center for the manufacturing industry.

\bibliographystyle{scsproc}
\bibliography{ms}

\begin{thebibliography}{}

\bibitem[\protect\citeauthoryear{{ANSYS Inc}}{{ANSYS Inc}}{2021}]{ansys}
{ANSYS Inc} 2021.
\newblock ``Ansys AVxcelerate Sensors Test and Validate Sensors for
  Self-Driving Cars''.
\newblock
  \url{https://www.ansys.com/de-de/products/av-simulation/ansys-avxcelerate-sensors}.
\newblock Accessed Sep. 10, 2021.

\bibitem[\protect\citeauthoryear{{Cosys-Lab}}{{Cosys-Lab}}{2022}]{Cosys-AirSimRepository}
{Cosys-Lab} 2022.
\newblock ``Cosys-AirSim Github Repository''.
\newblock \url{https://github.com/Cosys-Lab/Cosys-AirSim}.
\newblock Accessed Sep. 10, 2021.

\bibitem[\protect\citeauthoryear{{Cyberbotics Ltd.}}{{Cyberbotics
  Ltd.}}{2021}]{webots}
{Cyberbotics Ltd.} 2021.
\newblock ``Webots: Open-source Mobile Robot Simulation Software''.
\newblock \url{http://www.cyberbotics.com}.
\newblock Accessed Sep. 10, 2021.

\bibitem[\protect\citeauthoryear{Dannheim, Icking, Mader, and Sallis}{Dannheim
  et~al.}{2014}]{Dannheim2014WeatherSystems}
Dannheim, C., C.~Icking, M.~Mader, and P.~Sallis. 2014, 3.
\newblock ``{Weather detection in vehicles by means of camera and LIDAR
  systems}''.
\newblock {\em Proceedings - 6th International Conference on Computational
  Intelligence, Communication Systems and Networks, CICSyN 2014\/}, pp.
  186--191.


\bibitem[\protect\citeauthoryear{Dosovitskiy, Ros, Codevilla, Lopez, and
  Koltun}{Dosovitskiy et~al.}{2017}]{Dosovitskiy2017CARLA:Simulator}
Dosovitskiy, A., G.~Ros, F.~Codevilla, A.~Lopez, and V.~Koltun. 2017.
\newblock ``{CARLA}: {An} Open Urban Driving Simulator''.
\newblock In {\em Proceedings of the 1st Annual Conference on Robot Learning},
  pp.  1--16.

\bibitem[\protect\citeauthoryear{{dSPACE}}{{dSPACE}}{2021}]{aurelion}
{dSPACE} 2021.
\newblock ``{AURELION Lidar Model}''.
\newblock
  \url{https://www.dspace.com/fr/fra/home/products/sw/experimentandvisualization/aurelion_sensor-realistic_sim/aurelion_lidar.cfm}.
\newblock Accessed Sep. 10, 2021.

\bibitem[\protect\citeauthoryear{Geiger, Lenz, and Urtasun}{Geiger
  et~al.}{2012}]{geiger2012cvpr}
Geiger, A., P.~Lenz, and R.~Urtasun. 2012.
\newblock ``{Are we ready for Autonomous Driving? The KITTI Vision Benchmark
  Suite}''.
\newblock In {\em Proc.~of the IEEE Conf.~on Computer Vision and Pattern
  Recognition (CVPR)}, pp.  3354--3361.

\bibitem[\protect\citeauthoryear{Goodin, Carruth, Doude, and Hudson}{Goodin
  et~al.}{2019}]{Goodin2019PredictingADAS}
Goodin, C., D.~Carruth, M.~Doude, and C.~Hudson. 2019, 1.
\newblock ``{Predicting the influence of rain on LIDAR in ADAS}''.
\newblock {\em Electronics (Switzerland)\/}~vol. 8 (1).


\bibitem[\protect\citeauthoryear{Jansen, Huebel, and Steckel}{Jansen
  et~al.}{2022}]{lidarsim2022jansen}
Jansen, W., N.~Huebel, and J.~Steckel. 2022.
\newblock ``Physical LiDAR Simulation in Real-Time Engine''.
\newblock In {\em 2022 IEEE Sensors}, pp.  1--4.

\bibitem[\protect\citeauthoryear{Jhaldiyal and Chaudhary}{Jhaldiyal and
  Chaudhary}{2022}]{Jhaldiyal2022SemanticSO}
Jhaldiyal, A., and N.~Chaudhary. 2022.
\newblock ``Semantic segmentation of 3D LiDAR data using deep learning: a
  review of projection-based methods''.
\newblock {\em Applied Intelligence\/}.


\bibitem[\protect\citeauthoryear{{JPL} and {Caltech}}{{JPL} and
  {Caltech}}{2021}]{JetPropulsionLaboratoryECOSTRESS1.0}
{JPL} and {Caltech} 2021.
\newblock ``ECOSTRESS Spectral Library—Version 1.0''.
\newblock \url{https://ecostress.jpl.nasa.gov/}.
\newblock Accessed Sep. 10, 2021.

\bibitem[\protect\citeauthoryear{Koenig and Howard}{Koenig and
  Howard}{2004}]{Koenig2004DesignSimulator}
Koenig, N., and A.~Howard. 2004.
\newblock ``{Design and use paradigms for Gazebo, an open-source multi-robot
  simulator}''.
\newblock {\em 2004 IEEE/RSJ International Conference on Intelligent Robots and
  Systems (IROS)\/}~vol. 3, pp. 2149--2154.


\bibitem[\protect\citeauthoryear{Lewandowski, Eichinger, Kruger, and
  Krajewski}{Lewandowski et~al.}{2009}]{Lewandowski2009Lidar-basedEvidence}
Lewandowski, P.~A., W.~E. Eichinger, A.~Kruger, and W.~F. Krajewski. 2009.
\newblock ``{Lidar-based estimation of small-scale rainfall: Empirical
  evidence}''.
\newblock {\em Journal of Atmospheric and Oceanic Technology\/}~vol. 26 (3),
  pp. 656--664.


\bibitem[\protect\citeauthoryear{Li, Chen, and Lee}{Li et~al.}{2018}]{8579077}
Li, J., B.~M. Chen, and G.~H. Lee. 2018.
\newblock ``SO-Net: Self-Organizing Network for Point Cloud Analysis''.
\newblock In {\em 2018 IEEE/CVF Conference on Computer Vision and Pattern
  Recognition}, pp.  9397--9406.

\bibitem[\protect\citeauthoryear{Li, Bu, Sun, Wu, Di, and Chen}{Li
  et~al.}{2018}]{NEURIPS2018_f5f8590c}
Li, Y., R.~Bu, M.~Sun, W.~Wu, X.~Di, and B.~Chen. 2018.
\newblock ``PointCNN: Convolution On X-Transformed Points''.
\newblock In {\em Advances in Neural Information Processing Systems}, edited
  by\ S.~Bengio, H.~Wallach, H.~Larochelle, K.~Grauman, N.~Cesa-Bianchi, and
  R.~Garnett, Volume~31, Curran Associates, Inc.

\bibitem[\protect\citeauthoryear{Muckenhuber, Holzer, and Bockaj}{Muckenhuber
  et~al.}{2020}]{Muckenhuber2020AutomotiveCapabilities}
Muckenhuber, S., H.~Holzer, and Z.~Bockaj. 2020, 6.
\newblock ``{Automotive Lidar Modelling Approach Based on Material Properties
  and Lidar Capabilities}''.
\newblock {\em Sensors\/}~vol. 20 (11), pp. 3309.


\bibitem[\protect\citeauthoryear{{NVIDIA}}{{NVIDIA}}{2021}]{driveworks}
{NVIDIA} 2021.
\newblock ``DriveWorks SDK''.
\newblock \url{https://developer.nvidia.com/drive/driveworks}.
\newblock Accessed Sep. 10, 2021.

\bibitem[\protect\citeauthoryear{Qi, Yi, Su, and Guibas}{Qi
  et~al.}{2017}]{qi2017pointnetplusplus}
Qi, C.~R., L.~Yi, H.~Su, and L.~J. Guibas. 2017.
\newblock ``PointNet++: Deep Hierarchical Feature Learning on Point Sets in a
  Metric Space''.
\newblock {\em arXiv preprint arXiv:1706.02413\/}.


\bibitem[\protect\citeauthoryear{Qiu and Yuille}{Qiu and
  Yuille}{2016}]{QiuUnrealCV:Engine}
Qiu, W., and A.~Yuille. 2016.
\newblock ``UnrealCV: Connecting Computer Vision to Unreal Engine''.
\newblock In {\em Computer Vision -- ECCV 2016 Workshops}, edited by\ G.~Hua
  and H.~J{\'e}gou, pp.  909--916.
\newblock Cham, Springer International Publishing.

\bibitem[\protect\citeauthoryear{Qiu, Zhong, Zhang, Qiao, Xiao, Kim, Wang, and
  Yuille}{Qiu et~al.}{2017}]{Qiu2017UnrealCV:Vision}
Qiu, W., F.~Zhong, Y.~Zhang, S.~Qiao, Z.~Xiao, T.~S. Kim, Y.~Wang, and
  A.~Yuille. 2017.
\newblock ``UnrealCV: Virtual Worlds for Computer Vision''.
\newblock {\em ACM Multimedia Open Source Software Competition\/}.


\bibitem[\protect\citeauthoryear{Quigley, Gerkey, Conley, Faust, Foote, Leibs,
  Berger, Wheeler, and Ng}{Quigley et~al.}{2009}]{Quigley2009}
Quigley, M., B.~Gerkey, K.~Conley, J.~Faust, T.~Foote, J.~Leibs, E.~Berger,
  R.~Wheeler, and A.~Ng. 2009, 1.
\newblock ``ROS: an open-source Robot Operating System''.

\bibitem[\protect\citeauthoryear{{rFpro}}{{rFpro}}{2021}]{rfpro}
{rFpro} 2021.
\newblock ``{ADAS {\&} Autonomous}''.
\newblock \url{https://www.rfpro.com/virtual-test/adas-and-autonomous/}.
\newblock Accessed Sep. 10, 2021.

\bibitem[\protect\citeauthoryear{Rong, Shin, Tabatabaee, Lu, Lemke, Možeiko,
  Boise, Uhm, Gerow, Mehta, Agafonov, Kim, Sterner, Ushiroda, Reyes,
  Zelenkovsky, and Kim}{Rong et~al.}{2020}]{Rong2020LGSVLDriving}
Rong, G., B.~H. Shin, H.~Tabatabaee, Q.~Lu, S.~Lemke, M.~Možeiko, E.~Boise,
  G.~Uhm, M.~Gerow, S.~Mehta, E.~Agafonov, T.~H. Kim, E.~Sterner, K.~Ushiroda,
  M.~Reyes, D.~Zelenkovsky, and S.~Kim. 2020.
\newblock ``LGSVL Simulator: A High Fidelity Simulator for Autonomous
  Driving''.
\newblock In {\em 2020 IEEE 23rd International Conference on Intelligent
  Transportation Systems (ITSC)}, pp.  1--6.

\bibitem[\protect\citeauthoryear{Schouten, Jansen, and Steckel}{Schouten
  et~al.}{2021}]{Schouten2021SimulationSLAM}
Schouten, G., W.~Jansen, and J.~Steckel. 2021, 1.
\newblock ``{Simulation of Pulse-Echo Radar for Vehicle Control and SLAM}''.
\newblock {\em Sensors 2021\/}~vol. 21 (2), pp. 523.


\bibitem[\protect\citeauthoryear{Shah, Dey, Lovett, and Kapoor}{Shah
  et~al.}{2018}]{Shah2017AirSimVehicles}
Shah, S., D.~Dey, C.~Lovett, and A.~Kapoor. 2018.
\newblock ``{AirSim: High-Fidelity Visual and Physical Simulation for
  Autonomous Vehicles}''.
\newblock In {\em Field and Service Robotics}, edited by\ M.~Hutter and
  R.~Siegwart, pp.  621--635.
\newblock Cham, Springer International Publishing.

\bibitem[\protect\citeauthoryear{{Siemens}}{{Siemens}}{2021}]{siemensprescan}
{Siemens} 2021.
\newblock ``Lidar simulation and validation for autonomous vehicles''.
\newblock
  \url{https://www.plm.automation.siemens.com/global/en/webinar/lidar-simulation/87062}.
\newblock Accessed Sep. 10, 2021.

\bibitem[\protect\citeauthoryear{{The MathWorks Inc.}}{{The MathWorks
  Inc.}}{2021}]{mathworksautomation}
{The MathWorks Inc.} 2021.
\newblock ``{MATLAB: Automated Driving Toolbox}''.
\newblock \url{https://mathworks.com/products/automated-driving.html}.
\newblock Accessed Sep. 10, 2021.

\bibitem[\protect\citeauthoryear{{Vector Informatik}}{{Vector
  Informatik}}{2021}]{dyna4}
{Vector Informatik} 2021.
\newblock ``Vector and DYNA4: Environment Perception for ADAS and Autonomous
  Driving''.
\newblock
  \url{https://www.vector.com/int/en/products/products-a-z/software/dyna4/sensor-simulation/}.
\newblock Accessed Sep. 10, 2021.

\bibitem[\protect\citeauthoryear{{VIRES}}{{VIRES}}{2021}]{vtd}
{VIRES} 2021.
\newblock ``Vires and VTD, Sensor modeling in Autonomous Driving''.
\newblock \url{https://vires.mscsoftware.com/solutions/sensors/}.
\newblock Accessed Sep. 10, 2021.

\end{thebibliography}


\textbf{\uppercase{WOUTER JANSEN}} is a doctoral student at the Faculty of Applied Engineering at the University. Their email address is \email{wouter.jansen@uantwerpen.be}.

\textbf{\uppercase{ERIK VERREYCKEN}} is a doctoral student at the Faculty of Applied Engineering at the University
of Antwerp. Their email address is \email{erik.verreycken@uantwerpen.be}.

\textbf{\uppercase{ANTHONY SCHENCK}} is a doctoral student at the Faculty of Applied Engineering at the University
of Antwerp. Their email address is \email{anthony.schenck@uantwerpen.be}.

\textbf{\uppercase{JEAN-EDOUARD BLANQUART}} is a research engineer at Flanders Make. Their email address is \email{jeanedouard.blanquart@flandersmake.be}.

\textbf{\uppercase{CONNOR VERHULST}} is an associate application engineer at Flanders Make. Their email address is \email{connor.verhulst@flandersmake.be}.

\textbf{\uppercase{NICO HUEBEL}} is a researcher at the Faculty of Applied Engineering at the University
of Antwerp. Their email address is \email{nico.huebel@uantwerpen.be}.

\textbf{\uppercase{JAN STECKEL}} is a professor at the Faculty of Applied Engineering at the University
of Antwerp. Their email address is \email{jan.steckel@uantwerpen.be}.

\end{document}